%% file: main.tex
\documentclass[10pt,twocolumn,letterpaper]{article}

\usepackage{cvpr}
\usepackage{times}
\usepackage{epsfig}
\usepackage{graphicx}
\usepackage{amsmath}
\usepackage{amssymb}
\usepackage[numbers,sort&compress]{natbib}

\usepackage[pagebackref=true,breaklinks=true,letterpaper=true,colorlinks,bookmarks=false]{hyperref}

\newcommand{\Section}[1]{\vspace{-1mm} \section{#1} \vspace{-2mm}}
  \newcommand{\Subsection}[1]{\vspace{-1mm} \subsection{#1} \vspace{-1mm} }
  \newcommand{\Subsubsection}[1]{\vspace{0mm} \subsubsection{#1} \vspace{-2mm} }

\cvprfinalcopy 

\def\cvprPaperID{9} 

\ifcvprfinal\fi
\begin{document}

\title{The iWildCam 2021 Competition Dataset}

\maketitle

\input{abstract}
\input{introduction}

\input{data_preparation}
\input{evaluation}
\input{baseline_results}

\input{conclusion}
\ifcvprfinal
\input{acknowledgements}
\fi
{\small
\bibliographystyle{ieee_fullname}
\bibliography{main}
}




\end{document}

%% file: abstract.tex
\begin{abstract}
   Camera traps enable the automatic collection of large quantities of image data. Ecologists use camera traps to monitor animal populations all over the world. In order to estimate the abundance of a species from camera trap data, ecologists need to know not just which species were seen, but also how many individuals of each species were seen. Object detection techniques can be used to find the number of individuals in each image. However, since camera traps collect images in motion-triggered bursts, simply adding up the number of detections over all frames is likely to lead to an incorrect estimate. Overcoming these obstacles may require incorporating spatio-temporal reasoning or individual re-identification in addition to traditional species detection and classification.

   We have prepared a challenge where the training data and test data are from different cameras spread across the globe. The set of species seen in each camera overlap, but are not identical. The challenge is to classify species and count individual animals across sequences in the test cameras.
\end{abstract}

%% file: introduction.tex
\Section{Introduction}
The computer vision community has been making steady progress improving automated systems for species classification and localization in camera trap images over the past decade \cite{wilber2013animal,chen2014deep,zhang2016animal,miguel2016finding,giraldo2017camera,yousif2017fast,villa2017towards,norouzzadeh2017automatically, beery2018recognition, beery2020synthetic, beery2019long, schneider2018deep, beery2019efficient, tabak2020improving, norouzzadeh2019deep, beery2020context}. Classifications of species seen in a given image or sequence are used by ecologists to generate species richness models \cite{dorazio2006estimating}, species occurrence models \cite{mackenzie2017occupancy} or species distribution models \cite{elith2009species}, which describe (stated simply) where in a region or around the world a species might live (or be able to live). However, these types of models do not typically describe the \emph{abundance} (population size of a given species in an area) or \emph{density} (how that population is spatially distributed \cite{rowcliffe2013clarifying}) of the species. A common method for population estimation is \emph{mark-recapture}, which requires individual animals to be identified and recognized in future imagery \cite{silver2004use}. Though strides are being made in visual re-identification for species with strong biometric markings such as zebras \cite{schneider2019past, vidal2021perspectives}, many species are not visually re-identifiable by humans, making data collection and analysis difficult. To address this, ecological models have been developed that estimate abundance from counts of individuals of a species captured in each camera across short time windows \cite{moeller2018three,rowcliffe2013clarifying}. The iWildCam 2021 competition \footnote{iWildCam 2021 is hosted on Kaggle: \href{https://www.kaggle.com/c/iwildcam2021-fgvc8}{https://www.kaggle.com/c/iwildcam2021-fgvc8}} seeks to automate that counting process to enable abundance estimation to scale efficiently to large data collections, and one day to global data repositories such as Wildlife Insights \cite{ahumada2020wildlife}.
 
\begin{figure}
    \centering
    \includegraphics[width=0.45\textwidth]{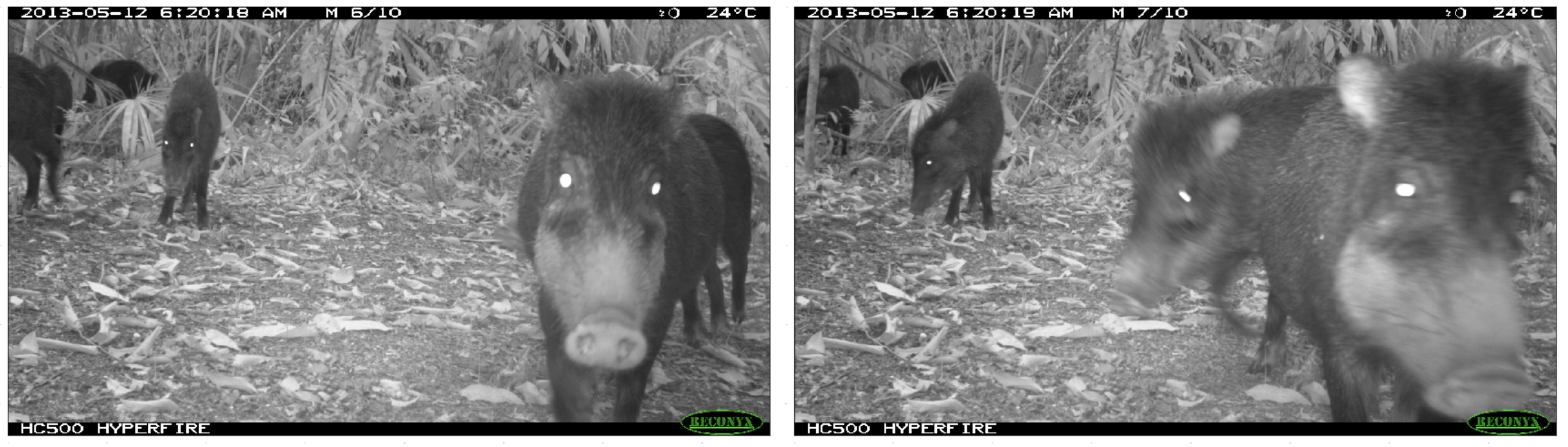}
    \caption{\textbf{How many pigs are there?} This year's challenge focuses on counting individuals across a sequence of camera trap images. Because the images are taken no faster than one frame per second, there are often temporal discontinuities between frames that make traditional tracking methods perform badly. However, humans are able to use a combination of spatio-temporal logic and visual re-identification to match individuals between frames.}
    \label{fig:splash}
\end{figure}

Competitors will categorize and count species across short bursts of images in the test data. No count labels have been provided for the training set, in hopes that competitors will develop methods that can learn to count without explicit training labels, as most  public camera trap data is not labeled with counts \cite{LILA}. We provide competitors with species labels along with weakly-supervised detections \cite{beery2019efficient} and instance segmentations \cite{birodkar2021surprising} to help them to disambiguate individuals. The competition also maintains the multi-modal aspects of the iWildcam 2020 challenge \cite{beery2020iwildcam} by providing citizen science images for the species of interest, remote sensing imagery for each camera location, and obfuscated geolocation for most cameras. 


%% file: data_preparation.tex
\Section{Data Preparation}
The dataset consists of three primary components: (i) camera trap images, (ii) citizen science images, and (iii) multispectral imagery for each camera location. Each component represents one technique for monitoring an ecosystem, each of which has unique strengths and limitations.
Species classification and localization performance has been shown to improve by using information beyond the image itself \cite{macaodha2019presence,chu2019geo,beery2020context} so we hope that participants will find creative and effective uses for this data and see similar improvements in species counting. 
 
\begin{figure}
\begin{minipage}[b]{.3\linewidth}
  \centering
  \centerline{\includegraphics[width=3cm]{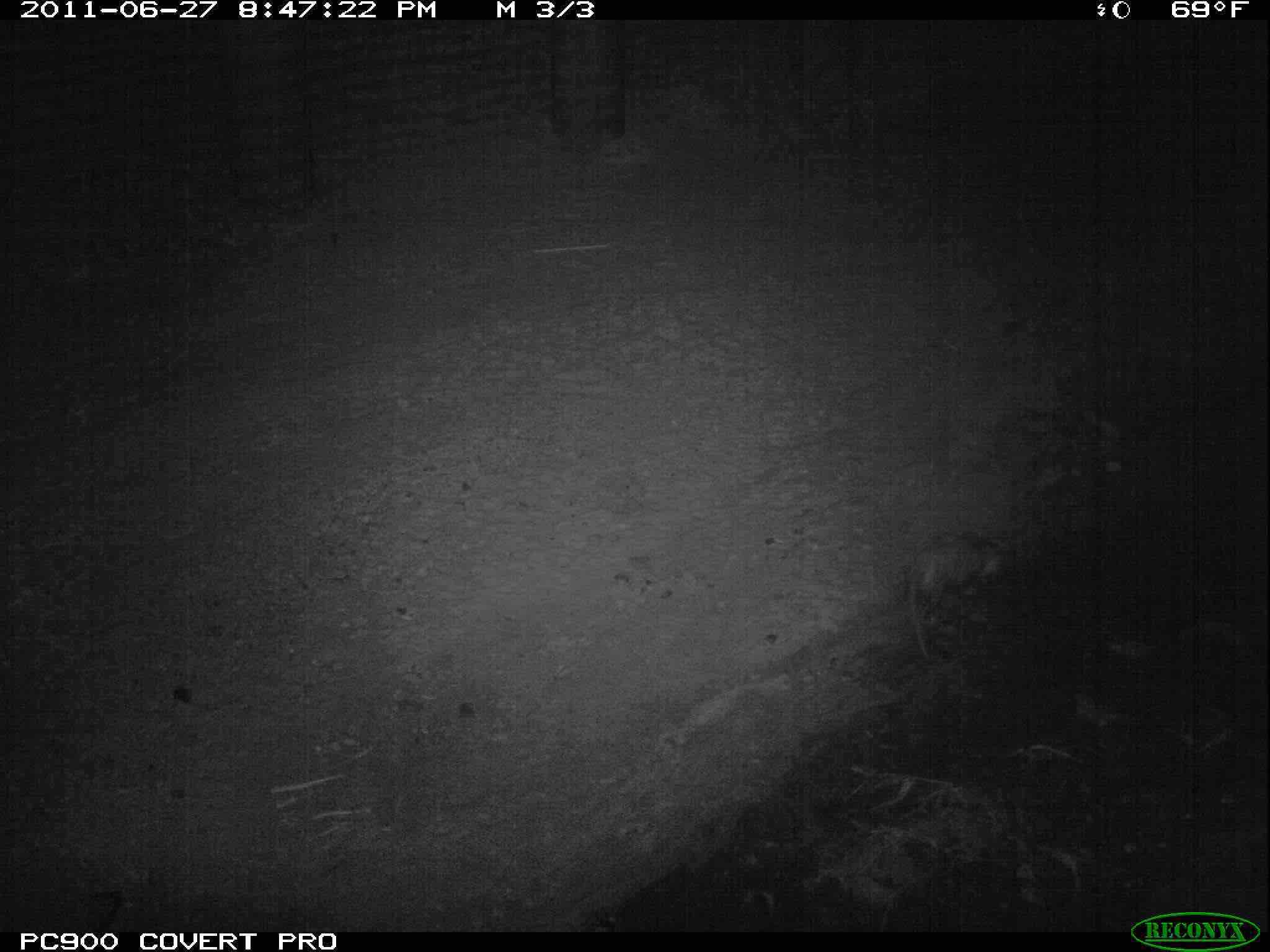}}
  \vspace{.05cm}
  \centerline{(1) Illumination}\medskip
\end{minipage}
\hfill
\begin{minipage}[b]{0.3\linewidth}
  \centering
  \centerline{\includegraphics[width=3cm]{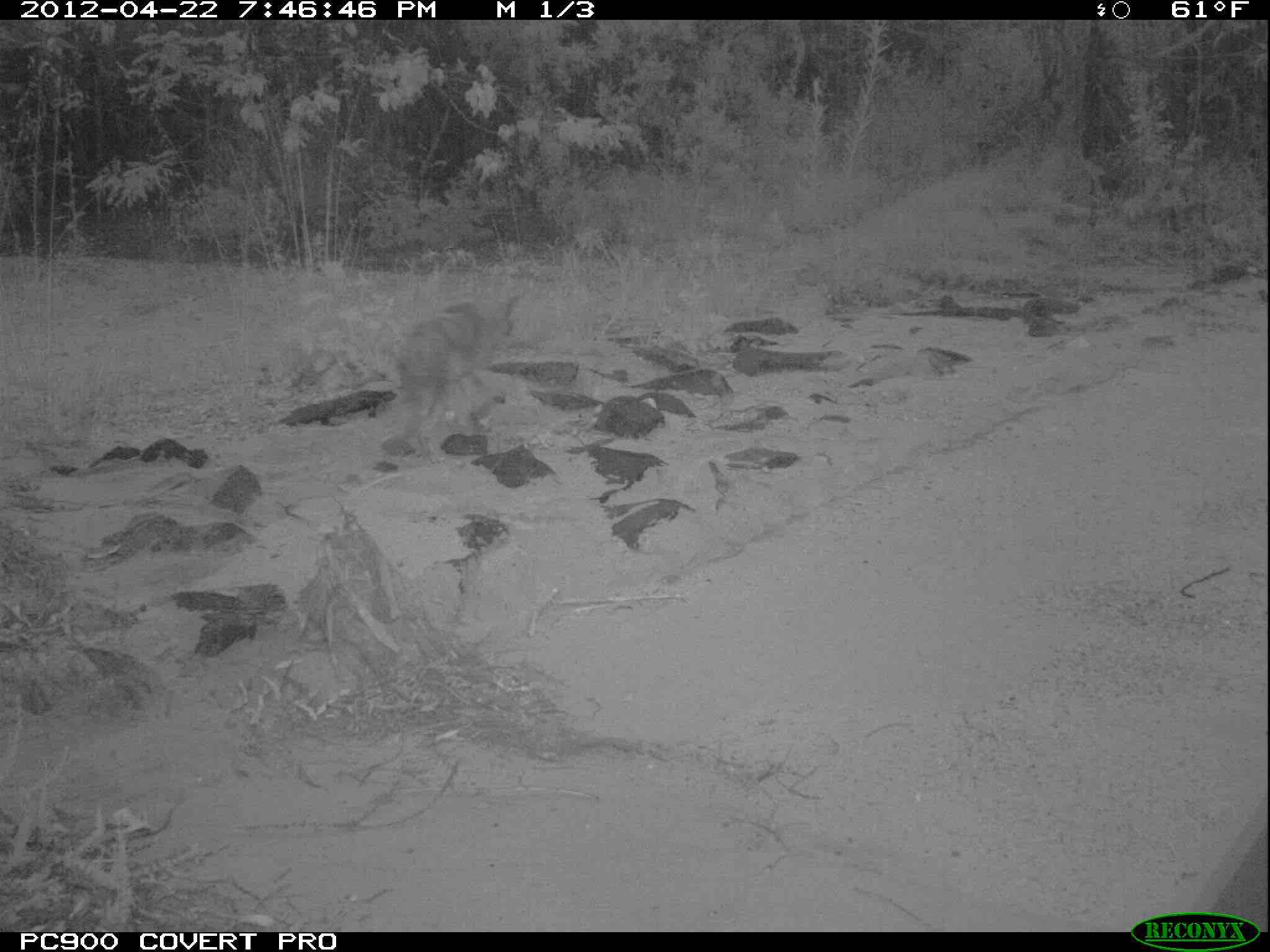}}
  \vspace{.05cm}
  \centerline{(2) Blur}\medskip
\end{minipage}
\hfill
\begin{minipage}[b]{.3\linewidth}
  \centering
  \centerline{\includegraphics[width=3cm]{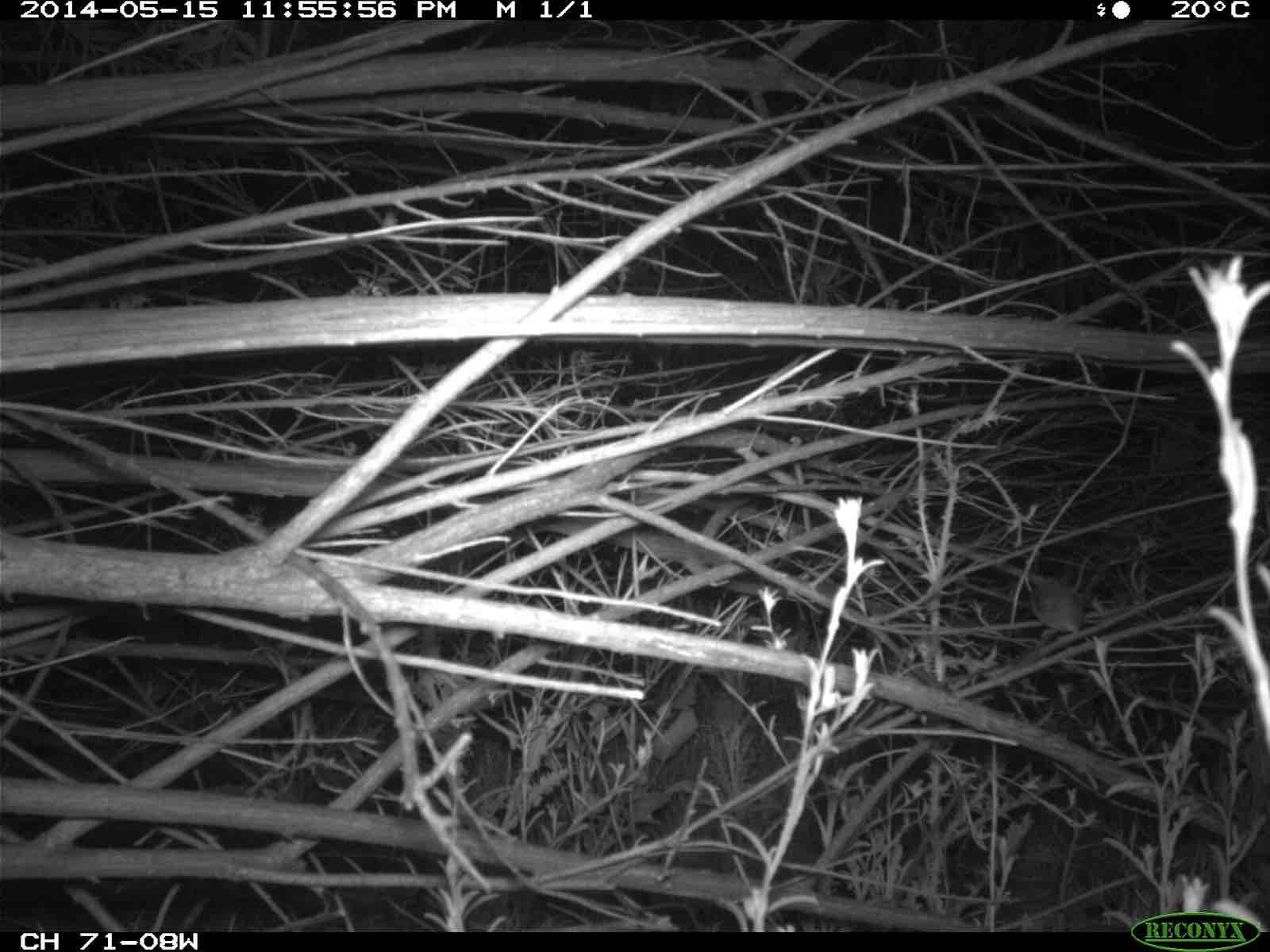}}
  \vspace{.05cm}
  \centerline{(3) ROI Size}\medskip
 \end{minipage}
\begin{minipage}[b]{0.3\linewidth}
  \centering
  \centerline{\includegraphics[width=3cm]{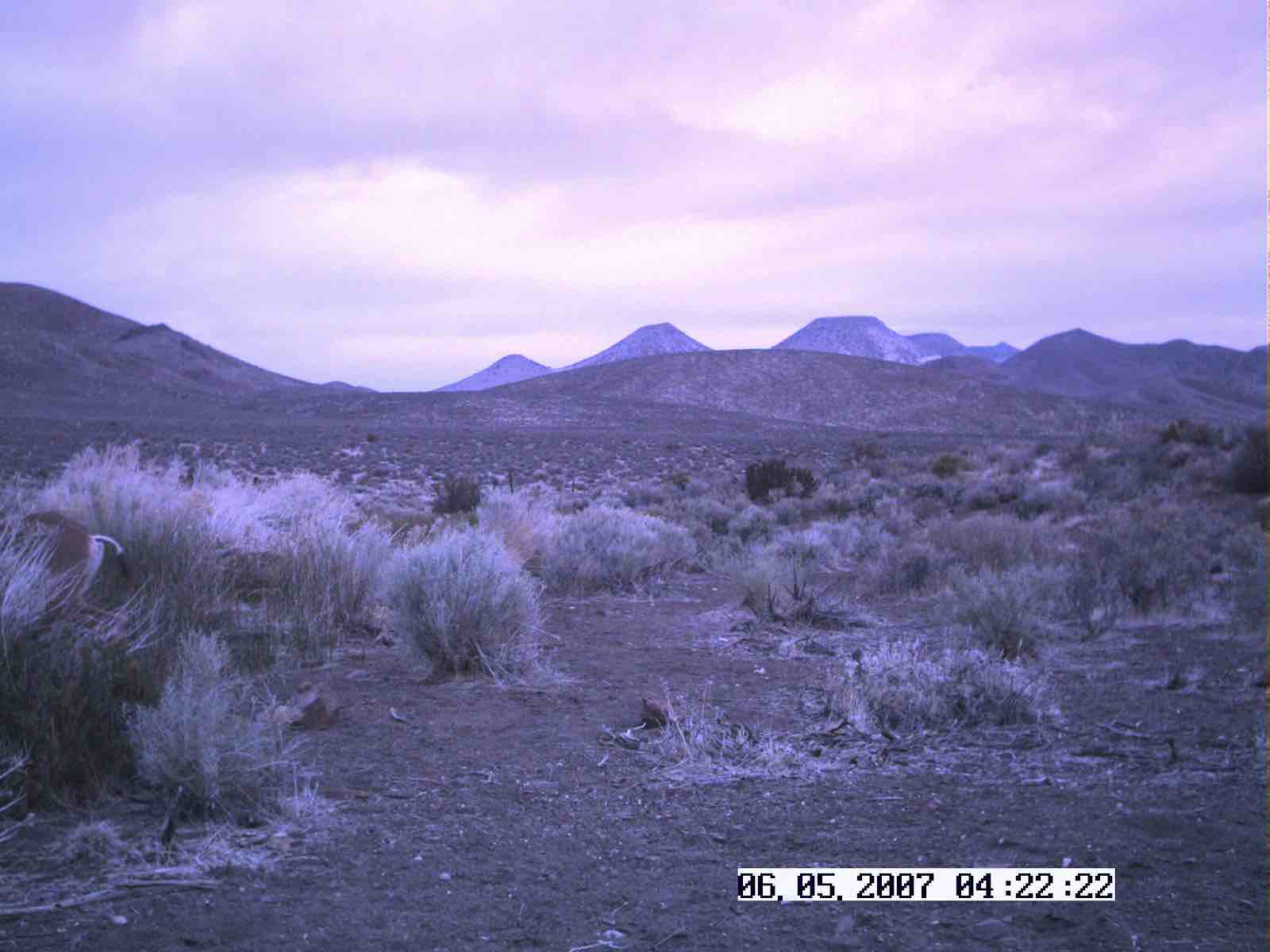}}
  \centerline{(4) Occlusion}\medskip
\end{minipage}
\hfill
\begin{minipage}[b]{.3\linewidth}
  \centering
  \centerline{\includegraphics[width=3cm]{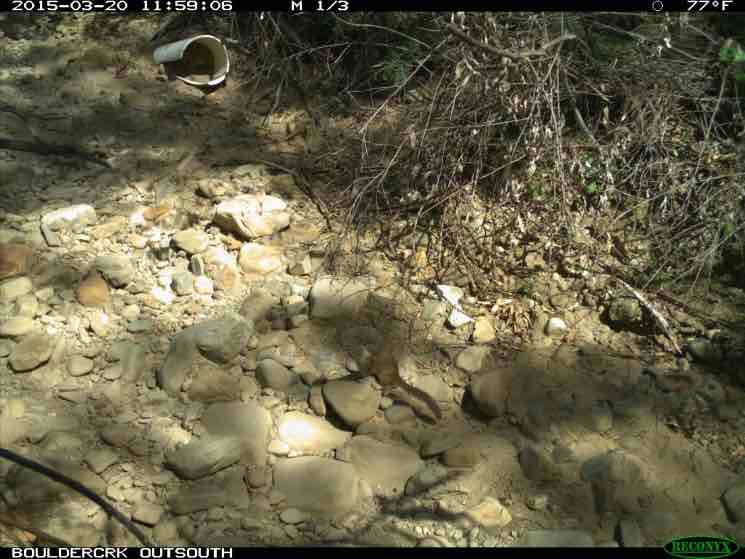}}
  \centerline{(5) Camouflage}\medskip
\end{minipage}
\hfill
\begin{minipage}[b]{0.3\linewidth}
  \centering
  \centerline{\includegraphics[width=3cm]{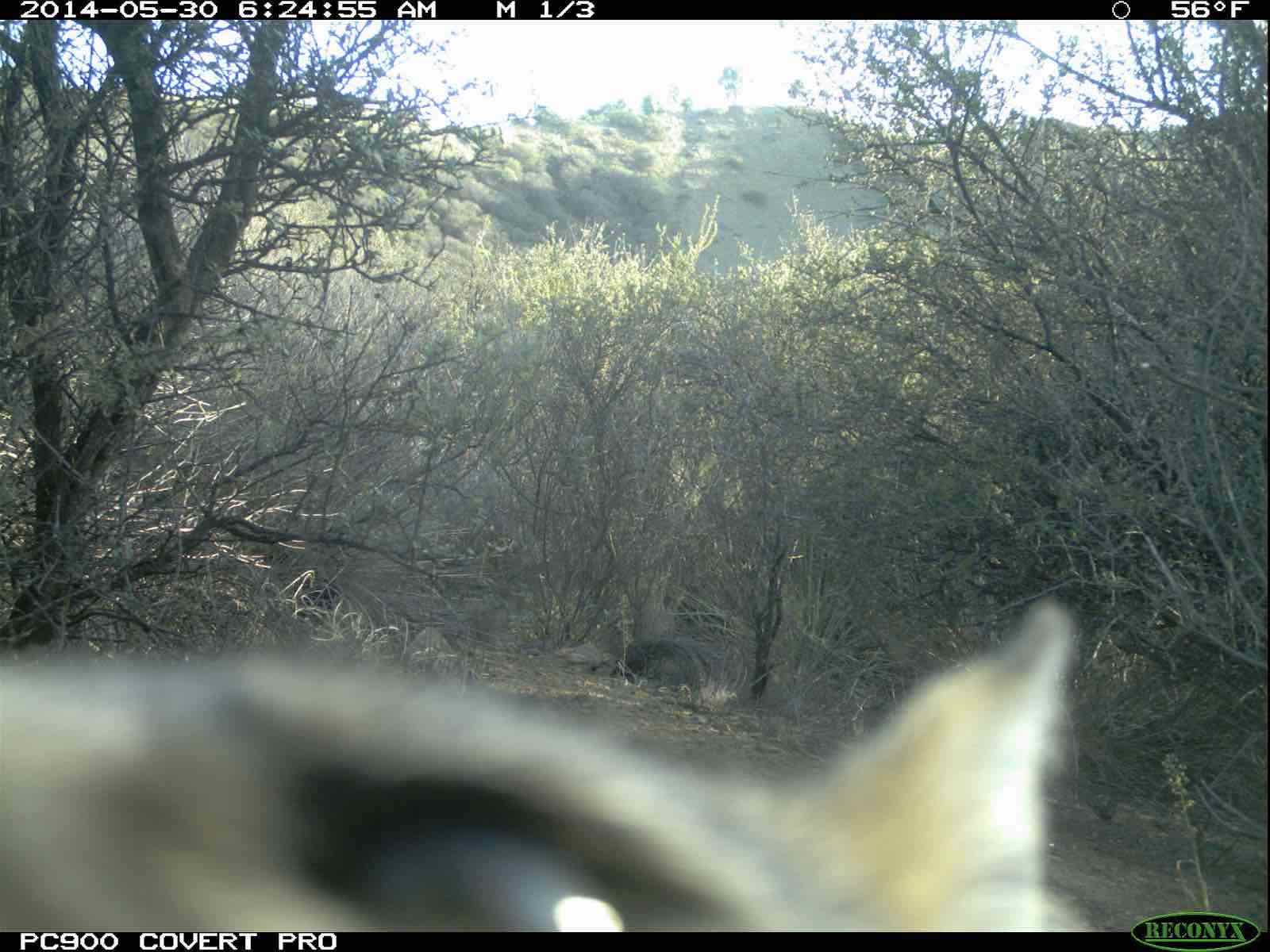}}
  \centerline{(6) Perspective}\medskip
\end{minipage}
\caption{\textbf{Common data challenges in camera trap images.} (1) {\bf Illumination}: Animals are not always well-lit. (2) {\bf Motion blur}: common with poor illumination at night. (3) {\bf Size of the region of interest} (ROI): Animals can be small or far from the camera. (4) {\bf Occlusion}: e.g. by bushes or rocks. (5) {\bf Camouflage}: decreases saliency in animals' natural habitat. (6) {\bf Perspective}: Animals can be close to the camera, resulting in partial, non-standard views.}
\label{fig:challenging_ims}
\end{figure}

\Subsection{Camera Trap Data}
The camera trap data (along with expert species annotations) is provided by the Wildlife Conservation Society (WCS) \cite{wcs_cam_traps}.  Camera trap images are taken automatically based on a motion-triggered sensor, so there is no guarantee that the animal will be centered, focused, well-lit, or at an appropriate scale (they can be either very close or very far from the camera, each causing its own problems). 
 See Fig. \ref{fig:challenging_ims} for examples of these challenges. 
 Empty images pose an additional challenge, as up to 70\% of the photos at any given location may be triggered by something other than an animal, such as wind in the trees.

If we wish to build systems that are trained to detect and classify animals and then deployed to new locations without further training, we must measure the ability of machine learning and computer vision to \emph{generalize to new environments}~\cite{beery2018recognition, tabak2020improving, koh2020wilds}. 
This is central to the 2018 \cite{beery2018iwildcam}, 2019 \cite{beery2019iwildcam}, 2020 \cite{beery2020iwildcam} and 2021 iWildCam challenges, all of which split train and test data by camera location, so no images from the test cameras are included in the training set to avoid overfitting to one set of backgrounds \cite{beery2018recognition}.

\Subsection{iWildCam 2021 Dataset}
The 2021 training set contains $203,314$ images from $323$ locations, and the WCS test set contains $60,214$ images from $91$ locations. These $414$ locations are spread across $12$ countries in different parts of the world.
Each image is associated with a location ID so that images from the same location can be linked. In some cases, WCS biologists placed multiple cameras at the same location. We denote this with a sub-location ID, which communicates that the background and hardware of the camera at these sub-locations is different, but the physical location is the same.
As is typical for camera traps, approximately 50\% of the total number of images are empty (this varies per location). The iWildCam 2021 dataset is slightly smaller than the iWildCam 2020 dataset. We removed images from iWildCam 2020 that were found to be corrupted, mislabeled, or labeled with ambiguous categories like `start'. 

\begin{figure}
\begin{minipage}[b]{\linewidth}
  \centering
  \centerline{\includegraphics[width=9cm]{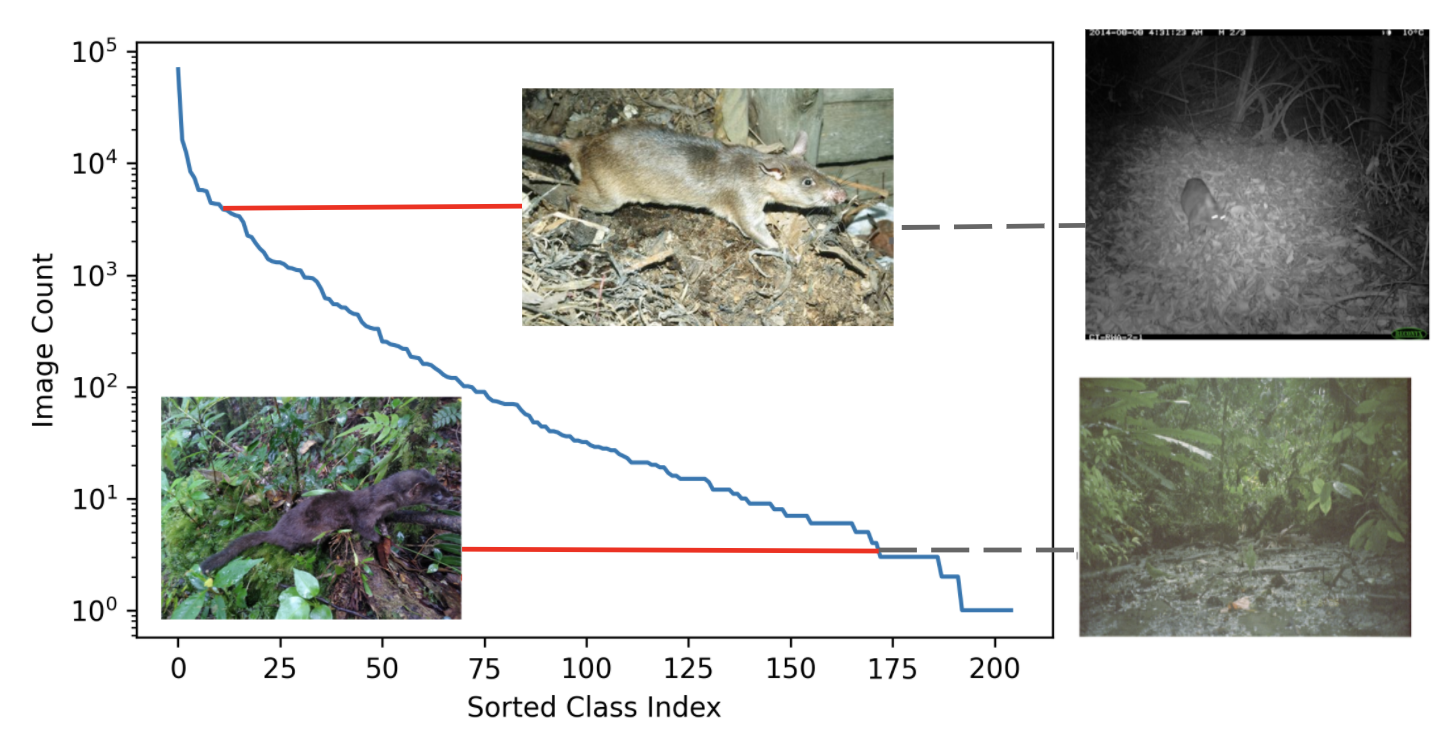}}
\end{minipage}

\caption{\textbf{Camera trap class distribution.} Per-class distribution of the camera trap data, which exhibits a long tail. We show examples of both a common class (the African giant pouched rat) and a rare class (the Indonesian mountain weasel). Within the plot we show images of each species, centered and focused, from iNaturalist. On the right we show images of each species within the frame of a camera trap, from WCS.}
\label{fig:camera_trap_distribution}
\end{figure}

There are $206$ species represented in the camera trap images.
The class distribution is long-tailed, as shown in Fig. \ref{fig:camera_trap_distribution}. 
Since we have split the data by location, some classes appear only in the training set.
Any images with classes that appeared only in the test set were removed. 

\Subsubsection{Count Labels}
Count labels for the test data were collected in collaboration with Centaur Labs \cite{centaurlabs}. We showed human annotators sequences of images that they could freely scroll through. Each sequence was labeled by between 3 and 30 individual annotators, with additional annotations collected for examples where annotators did not agree. Final counts were determined by majority vote, weighted by annotator performance on an expert-labeled subset. Sequences found to have multiple species were manually annotated by experts.

\Subsubsection{Obfuscated GPS Locations}
In order to allow competitors to try to use the geographic location of the cameras to improve their classification \cite{macaodha2019presence}, we worked with WCS to release obfuscated GPS coordinates for most of the camera trap locations. The precise coordinates of the cameras have been obfuscated randomly to within 1 km for privacy and security reasons, and correspond to the centers of the provided remote sensing imagery. Some of the obfuscated GPS locations were not released at the request of WCS, but we can confirm that all locations without GPS are from the same country.

\Subsection{iNaturalist Data}
iNaturalist is an online community where citizen scientists post photos of plants and animals and collaboratively identify the species \cite{inat}. Similar to iWildCam 2020, we provide a mapping from our classes into the iNaturalist taxonomy.\footnote{Note that for the purposes of the competition, competitors may only use iNaturalist data from the 2017-2021 iNaturalist competition datasets.}
We also provide the subsets of the iNaturalist 2017-2019 competition datasets \cite{van2018inaturalist} that correspond to species seen in the camera trap data.
This curated set provides $13,051$ additional images for training, covering $75$ classes. 

Though small relative to the camera trap data, the iNaturalist data has some unique characteristics.
First, the class distribution is completely different (though it is still long tailed).
Second, iNaturalist images are typically higher quality than the corresponding camera trap images, providing valuable examples for hard classes. 
See \cite{beery2020iwildcam} for a comparison between iNaturalist images and camera trap images.



\Subsection{Remote Sensing Data}
In addition to the raw remote sensing data for each camera location outlined in \cite{beery2020iwildcam}, this year we have provided pre-extracted ImageNet \cite{imagenet_cvpr09} features. 
We use an ImageNet-pretrained ResNet-50 \cite{he2016deep} to extract features from the RGB channels of each multispectral image. 

\Subsection{Provided Models}
\Subsubsection{The MegaDetector}
Competitors are free to use the Microsoft AI for Earth MegaDetector \cite{beery2019efficient} (a general and robust camera trap detection model )as they see fit. Megadetector V3 detects animal and human classes, while the MegaDetector V4 adds a vehicle class. Any version of the MegaDetector is allowed to be used in this competition. The models can be downloaded on the Microsoft Camera Traps GitHub repository \cite{microsoftcameratraps}. We provide the top MegaDetector V3 boxes and associated confidences along with our WCS image metadata.

\begin{figure}
\begin{minipage}[b]{\linewidth}
  \centering
  \centerline{\includegraphics[width=9cm]{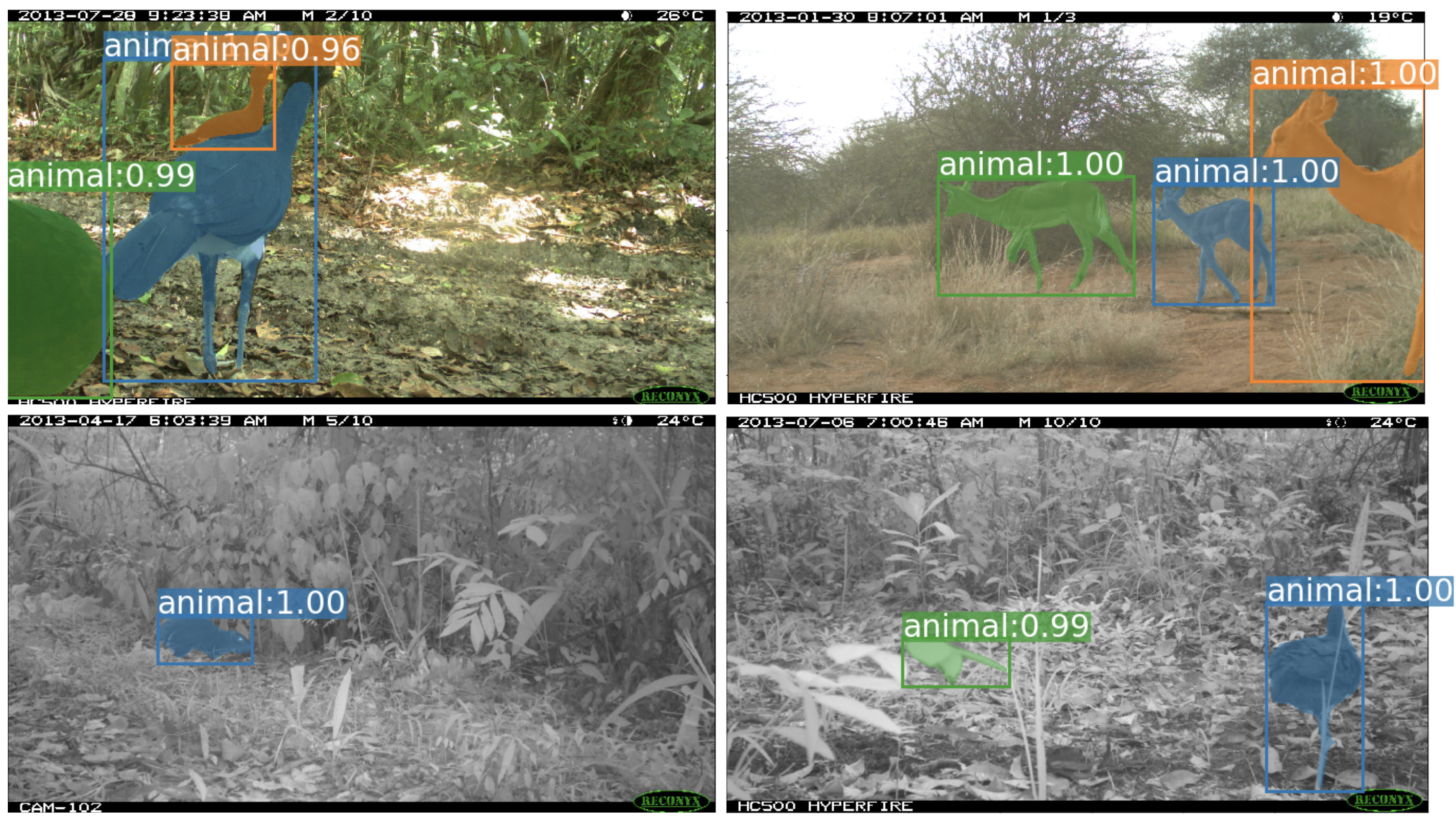}}
\end{minipage}
\label{fig:segs}
\caption{Segmentation results from DeepMAC, paired with MegaDetector V3 boxes. You can see in the lower right example that if the boxes are in error, the segmentation model will still provide its best guess at a segmentation (here it has segmented part of a plant that was a MegaDetector false positive).}
\end{figure}

\Subsubsection{DeepMAC}
Along with MegaDetector box labels, we also provide a method to extract corresponding segmentation masks within each detected box.
The segmentations are derived from the DeepMAC model \cite{birodkar2021surprising}. 
Although DeepMAC is designed as an instance segmentation model (i.e. detection+segmentation),
for this competition we provide an instance of the model which takes boxes as input from the user.
Combined with the MegaDetector box labels, or a user-provided detection model, this can be used to extract a per-detection
segmentation mask. We provide the DeepMAC masks associated with MegaDetector V3 boxes on Kaggle. Examples of segmentation results paired with MegaDetector V3 boxes can be seen in Fig. \ref{fig:segs}. The DeepMAC model was originally trained on all of COCO \cite{lin2014microsoft} and achieves a 
detection and mask mAP of 44.5 \% and 39.7 \% respectively.

%% file: evaluation.tex
\Section{Evaluation}
\begin{figure*}[h!]
    \centering
    \includegraphics[width=\linewidth]{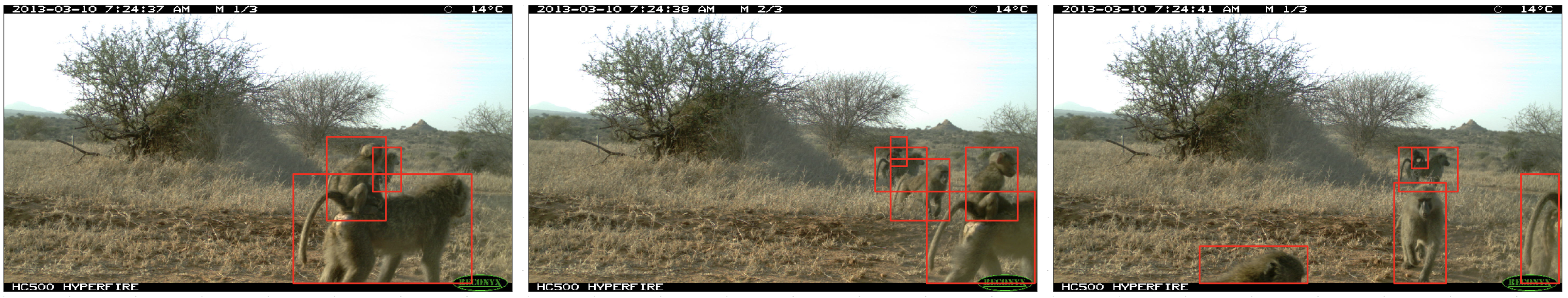}
    \caption{Here, the MegaDetector correctly boxed all animals and the classification model also correctly predected ``baboon" as the class for all three images in the sequence. Our majority vote classification for the sequence is therefore ``baboon" (correct) and our baseline model would see 5 boxes in both the second and 3rd image (the maximum number of boxes in any frame across the sequence) and predict ``5 baboons". This prediction is close, but in fact there is one baboon in image 2 that is not visible in image 3, and one baboon in image 3 that is new, so the correct answer for this sequence would be ``6 baboons".}
    \label{fig:baseline}
\end{figure*}

Let $X \in \{0, 1, 2, \ldots\}^{n \times m}$ be a matrix of predictions, so each entry $x_{ij}$ is the predicted count for species $j \in \{1,\ldots, m\}$ in sequence $i \in\{1, \ldots, n\}$. 
Let $Y\in \{0, 1, 2, \ldots\}^{n \times m}$ be the matrix of corresponding ground truth counts.
Submissions will be evaluated using \emph{mean columnwise root-mean-squared error} (MCRMSE) given by 
\begin{equation} \label{mcrmse}
    \textrm{MCRMSE}(X, Y) = \frac{1}{m} \sum_{j=1}^{m} \sqrt{\frac{1}{n} \sum_{i=1}^n (x_{ij} - y_{ij})^2}.
\end{equation}
We selected this metric out of the options provided by Kaggle in order to capture both species identification mistakes and count mistakes as well as to ensure false predictions on empty sequences would contribute to the error. Because many sequences are empty in camera trap data and because many species are rare, the metric tends to be a small number even when the actual errors in counts are large. To convert the metric to something more interpretable, we can un-normalize the metric from MCRMSE to the \emph{summed columnwise root summed squared error} (SCRSSE) given by
\begin{align} 
   \textrm{SCRSSE}(X,Y) &= m \sqrt{n} \textrm{MCRMSE}(X,Y) \nonumber \\
   &= \sum_{j=1}^{m} \sqrt{ \sum_{i=1}^n (x_{ij} - y_{ij})^2}. \label{scrsse}
\end{align}

%% file: baseline_results.tex
\Section{Baseline Results}

We built our simple counting baselines from our iWildCam 2020 classification baseline model (see details in \cite{beery2020iwildcam}),  the iWildCam 2020 winning submission, and the provided MegaDetector V3 results. The results can be seen in Table \ref{tab:results}, and the simple baselines are described below.
\begin{itemize}
\setlength\itemsep{0em}
    \item \textbf{Max boxes:}. We assume that all high-confidence animal boxes ($\geq0.8$) for an image are correct, and that the species in all boxes match our majority-vote classification prediction for that sequence. We take the maximum number of boxes from any image in the sequence and use that as our count. This will be a lower bound on the actual number of individuals across the sequence since it prevents double counting multiple images of the same individual. Example in Fig \ref{fig:baseline}.
    \item \textbf{Sum boxes:} We assume that all high-confidence animal boxes ($\geq0.8$) for each image are correct, and that the species in all boxes match our majority-vote classification prediction for that sequence. We take the sum of boxes across the sequence and use that as our count. This will be a upper bound on the actual number of individuals since individuals seen in multiple frames will be double counted.
    \item \textbf{One per predicted species:} We add a count of one for each unique species predicted by our image-level classification model across the sequence. This will be a lower bound on the actual number of individuals across the sequence as it just assumes that one animal was seen per species, regardless of detection results.
    \item \textbf{All zeros:} Just predict zero for all instances. Under our chosen metric this performs surprisingly well. This is for two reasons. First, camera trap data frequently has a small number of animals for any given species. Second, the model is double penalized if the count is correct but the species is incorrect (one penalty for missing the correct species count and one for overpredicting the incorrect species count).
\end{itemize}

\begin{table}[]
    \centering
    \begin{tabular}{|l|l|l|}
    \hline
 \textbf{Baseline} & \textbf{MCRMSE} & \textbf{SCRSSE}\\
 \hline 
 All zeros & 0.03938 & 844.73 \\
 \hline
 Max boxes A   & 0.05890 & 1263.50\\
 Sum boxes A   & 0.17550 & 3753.49\\
 One per predicted species A & 0.04061& 871.15\\
 \hline
 Max boxes B   & 0.03720 & 798.051\\
 Sum boxes B   & 0.19897 & 4268.13\\
 One per predicted species B & \textbf{0.03593}& \textbf{770.72}\\
 \hline
\end{tabular}
\vspace{5pt}
    \caption{\textbf{Simple baseline results on the test set.} For the (A) set, we used the classification predictions from our naive classification baseline from the iWildCam 2020 competition \cite{beery2020iwildcam}. For the (B) set we used the classification predictions from the iWildCam 2020 competition winning solution from Megvii Research Nanjing. We use the MegaDetector V3 boxes and a set of simple heuristics to generate counts from the species prediction.}
    \label{tab:results}
\end{table}

%% file: conclusion.tex
\Section{Conclusion}
The iWildCam 2021 dataset presents a new challenge for computer vision: counting the number of individuals across low-frame-rate sequences of images \cite{beery2018iwildcam,beery2019iwildcam,beery2020iwildcam}.
In subsequent years, we plan to extend the iWildCam challenge by adding additional data streams and tasks, such as detection, segmentation, or distance estimation. We hope to use the knowledge we gain throughout these challenges to facilitate the development of systems that can accurately provide real-time species ID and counts in camera trap images at a global scale. Any forward progress made will have a direct impact on the scalability of biodiversity research geographically, temporally, and taxonomically.

%% file: acknowledgements.tex
\Section{Acknowledgements}
We would like to thank Dan Morris and Siyu Yang (Microsoft AI for Earth) for their help curating the dataset, providing bounding boxes from the \href{https://github.com/microsoft/CameraTraps/blob/master/megadetector.md}{MegaDetector}, and hosting the data on Azure. We would like to thank Jonathan Huang and the Visual Dynamics Team at Google Research for providing segmentation labels. We thank the Wildlife Conservation Society for providing the camera trap data and species annotations, and Centaur Labs for working with us to label counts on the test set. We thank Kaggle for supporting the iWildCam competition for the past four years. Thanks also to the FGVC Workshop, Visipedia, and our advisor Pietro Perona for continued support. This work was supported in part by NSF GRFP Grant No. 1745301. The views are those of the authors and do not necessarily reflect the views of the NSF.